%% file: main.tex
\documentclass[letterpaper]{article} 
\usepackage[preprint]{aaai2027}  
\usepackage[hyphens]{url}  
\usepackage{graphicx} 
\usepackage{natbib}  
\usepackage{caption} 
\usepackage{algorithm}
\usepackage{algorithmic}

\usepackage{newfloat}
\usepackage{listings}
\DeclareCaptionStyle{ruled}{labelfont=normalfont,labelsep=colon,strut=off} 
\floatstyle{ruled}
\newfloat{listing}{tb}{lst}{}
\floatname{listing}{Listing}

\usepackage{booktabs}

\newcommand{\htv}{H$^2$V}
\newcommand{\htvv}{\mbox{H$^2$V-M}}

\usepackage{amsthm}
\theoremstyle{remark}

\nocopyright

\title{Validating, Not Sampling: Region-Level Robustness of Vision-Language and Vision-Language-Action Models}
\author{
Bogdan Aron\textsuperscript{\rm 1}\equalcontrib\footnote{This work was conducted while Bogdan Aron was interning at Safe Intelligence}, Christopher Brix\textsuperscript{\rm 2}\equalcontrib, Benedikt Brückner\textsuperscript{\rm 2}, Yanghao Zhang\textsuperscript{\rm 2}, \\
Panagiotis Kouvaros\textsuperscript{\rm 2}, Alessio Lomuscio\textsuperscript{\rm 2}
}
\affiliations{
\textsuperscript{\rm 1} Imperial College London \\
\textsuperscript{\rm 2} Safe Intelligence \\
\texttt{bogdan.aron23@imperial.ac.uk}, \texttt{\{christopher, benedikt, yanghao, panagiotis, alessio\}@safeintelligence.ai}
}

\begin{document}

\maketitle

\begin{abstract}
Vision-language models (VLMs) and vision-language-action models (VLAs) are increasingly deployed in real-world applications.
There, a small perturbation to the recorded camera image may change a decision significantly.
However, existing benchmarks for these models only sample perturbations, which does not guarantee the absence of a failure in the untested region.
We present the first robustness validation of six VLMs (drawn from the Gemma, InternVL, LLaVA, and Qwen families) and five VLAs (drawn from the GR00T, OpenVLA, and $\pi$ families) over entire continuous regions of photometric and geometric image perturbation: brightness shifts, camera rotations, and their composition.
To this end, we build on the validation framework \htv{} and introduce \htvv{}, a margin-aware convergence rule that makes validation affordable at the 32B parameter scale.
We demonstrate that \htvv{} outperforms \htv{} by an order of magnitude in model queries and that it finds counterexamples faster than random sampling while providing soundness guarantees.
Our VLM and VLA robustness validation shows that robustness is mostly dependent on the perturbation type, rather than the model, and that VLMs are more robust to large camera rotations than VLAs.
For VLAs, even perturbations as small as $\pm1^\circ$ can change the commanded action in many cases.
We also show that robustness depends more on model family than on model size.
\end{abstract}


\input{sections/introduction}
\input{sections/related_work}
\input{sections/h2v_intro}
\input{sections/general_changes_to_h2v}
\input{sections/vlm_setup}

\input{sections/vla_setup}
\input{sections/results}
\input{sections/conclusion}

%
{\small
\bibliography{bib_short}
}

\clearpage
\appendix
\input{sections/appendix}


\end{document}

%% file: sections/introduction.tex
\section{Introduction}

Vision-language models (VLMs) and vision-language-action models (VLAs) connect perception to language, planning, and action~\citep{Bai+25,Zitkovich+23}.
Their robustness is therefore not only a question of classification stability: a small visual change can alter a multiple-choice answer or the action a robot is commanded to execute.
Throughout, an \emph{adversarial example} is a bounded, meaning-preserving perturbation that changes the output in a way the specification forbids.
Here the bounded set is photometric, geometric, or both: a brightness shift, a camera rotation, or their composition, applied to the image inputs alone.
Such an example is therefore a physically plausible change to a camera image, whether caused by natural variation or by an attacker.

Robustness for these models is currently measured by sampling.
A benchmark applies a collection of corruptions, random perturbations, or manually designed attacks, and reports how often a model changes its answer or fails a downstream task~\citep{Qiu+25,Wang+25c}.
Such a protocol can exhibit a failure, but it cannot establish the absence of one: if no adversarial example is found, the result may simply mean that the sampled perturbations did not include one.
The limitation is well documented in the adversarial robustness literature, where imperceptible input changes flip predictions~\citep{Szegedy+14,GoodfellowShlensSzegedy15} and defences that appear robust under fixed or sampled attacks are repeatedly broken by stronger adaptive attacks~\citep{AthalyeCarliniWagner18,Tramer+20}, so a sound evaluation must reason over all perturbations in the bounded region rather than a sampled subset~\citep{Carlini+19}.
Its practical consequence has not been measured for VLMs and VLAs, because measuring it requires the ground truth that sampling cannot supply.

We supply it.
We build on \htv{}~\citep{ZhangKouvarosLomuscio25}, which maps a bounded, multidimensional perturbation region to a one-dimensional Hilbert curve and searches for a counterexample by H\"older optimisation.
Although \htv{} also queries the model at finitely many perturbations, it is not a sampling method in the above sense: every query tightens a H\"older lower bound on the objective that covers the entire region, including the uncountably many perturbations that are never evaluated.
A positive \htv{} outcome therefore rules out counterexamples everywhere in the region, provided the estimated H\"older constants upper-bound the true variation of the objective.
Applying it here is also a question of scale: the original evaluation validated vision models of up to 300M parameters, where each query is a single forward pass of a classifier, whereas our models reach 32B parameters and each query runs a full multimodal pipeline.
At this cost the fixed convergence threshold of \htv{} is unaffordable, so we introduce \htvv{}, a margin-aware rule that terminates as soon as the accumulated H\"older evidence suffices to validate the instance.

The resulting evaluation covers six VLMs and five VLAs over $57.4$M model queries, and yields conclusions that the sampled studies it replaces are structurally unable to reach.
\begin{itemize}
\item \textbf{We can bound the robust region from both sides.}
For the median LIBERO frame, every camera rotation up to $\pm1^\circ$ leaves the commanded action unchanged, and some rotation within $\pm2^\circ$ changes it.
Brightness is the opposite: no frame fails below a $\pm2\%$ shift, and the median frame tolerates $\pm30\%$.
Since a knocked or remounted camera is easily one or two degrees out, these models are fragile to a disturbance they will meet in normal operation, and robust to lighting changes far larger than a real sensor drifts.
An attack-based study could have missed the failing instances.
Only validation can show that the whole brightness range is clear.
\item \textbf{Sampled evaluation mislabels robustness at every affordable budget.}
A $k$-draw protocol wrongly calls an instance robust in $50\%$ of the cases where it reports robustness at $k=1$, $42\%$ at $k=10$, and $6\%$ at $k=100$.
Raising the budget a further hundredfold, to $17\,000$ draws per query, removes only a factor of three of that error and costs $11\times$ what validating the same instances does, and on $34$ instances that budget does not find the adversarial example at all.
\item \textbf{Robustness is a property of the task and perturbation, not of the model.}
The biggest difference in robustness is between the VLM and VLA families, with VLMs tolerating larger camera rotations than VLAs.
Within each family, the model size is less important than the choice of architecture: increasing the number of parameters does not make the Qwen2.5 model significantly more robust.

\end{itemize}

%% file: sections/related_work.tex
\section{Related Work}

\paragraph{Robustness validation and verification.}
Neural network verification methods aim to reason over dense input regions rather than finite perturbation samples: mixed-integer programming~\citep{TjengXiaoTedrake19}, SMT solving~\citep{Katz+17}, abstract interpretation~\citep{Gehr+18a,Singh+19a}, and branch-and-bound with bound propagation~\citep{Bunel+20}.
These provide strong formal guarantees when their assumptions and implementations hold, but even the complete verifiers that win the annual verification competition VNN-COMP~\citep{Xu+21,Kaulen+25} are designed for moderate-size networks and scale poorly to the large, black-box, multimodal models considered here~\citep{Liu+20a}.
\htv{} takes a different route, validating local robustness against low-dimensional image transformations using a Hilbert curve reduction and H\"older optimisation~\citep{ZhangKouvarosLomuscio25}; its soundness depends on the quality of the H\"older constant estimates.

\paragraph{Geometric and photometric robustness.}
Geometric and semantic robustness verification studies transformations such as rotation, translation, scaling, and brightness~\citep{Balunovic+19a,Mohapatra+20}.
Randomised smoothing yields probabilistic certificates~\citep{CohenRosenfeldKolter19}, including for parameterised image transformations~\citep{Li+21,FischerBaaderVechev20}, where bounds on the interpolation error extend the guarantee across an entire continuous parameter interval; those certificates hold with high probability over the smoothing samples and apply to the smoothed classifier rather than the deployed model.
Sampling-based attacks can instead reveal sensitivity to these transformations but cannot establish its absence, and for spatial transformations even strong first-order attacks fail to reliably find worst-case perturbations~\citep{Engstrom+19}.

\paragraph{VLM and VLA evaluation.}
Recent model families, including Qwen-VL~\citep{Bai+25,Bai+25b}, LLaVA~\citep{Liu+24b}, InternVL~\citep{Wang+25b}, and Gemma~\citep{Gemma+25}, have pushed VLM performance towards larger and more capable multimodal instruction followers, and benchmarks such as MMBench~\citep{Liu+24} evaluate their visual-question-answering competence using multiple-choice tasks.
Robustness studies for VLMs apply image corruptions~\citep{Qiu+25}, adversarial image perturbations~\citep{Zhang+25c,Zhao+23}, or multimodal jailbreak prompts~\citep{Qi+24}, and measure behaviour over the tested inputs; randomised smoothing has also been extended to probabilistic certificates for generative VLM outputs through an oracle classification task~\citep{Seferis+25}.
VLAs such as OpenVLA~\citep{Kim+24}, $\pi_0$-style action models~\citep{Black+25b,Black+25}, and GR00T~\citep{Bjorck+25} connect visual-language understanding to robot actions, and are harder to evaluate because action outputs can be continuous, tokenised, or chunked over a horizon.
Existing VLA studies evaluate concrete adversarial patches, backdoor triggers, positional changes, or sampled physical variations rather than validating every point in a bounded continuous region~\citep{Wang+25c,Zhou+25c,Pang+25}.
We focus on the open-loop one-step setting without internal reasoning; closed-loop simulation and models with internal chain-of-thought reasoning are left for future work.

\paragraph{What the sampled protocol cannot conclude.}
The VLM and VLA robustness studies above share a common structure: they report an output-change rate over a tested set of perturbations.
Three questions therefore lie outside their reach, and they are the questions this paper answers.
They cannot report a \emph{validated radius}, because a radius at which nothing was found is not a radius at which nothing exists, so they bound the robustness boundary from one side only.
They cannot report their own \emph{error rate}, because doing so requires knowing which of their apparently-robust instances actually admit a counterexample.
And they cannot establish that a \emph{model ranking} is a property of the models rather than of the attack, since an output-change rate measures model fragility and attack effectiveness together, whereas a positive validation outcome covers every perturbation in the region and so isolates the model.

%% file: sections/h2v_intro.tex
\section{Background: \htv{}}

\htv{} is a black-box method for validating the local robustness of neural networks against low-dimensional image transformations~\citep{ZhangKouvarosLomuscio25}.
Given a fixed input and a bounded $N$-dimensional transformation region $\Theta$, it expresses the robustness specification as the minimisation of a scalar objective $f(\theta)$ over all transformation parameters $\theta\in\Theta$, constructed so that $f(\theta)>0$ means the specification holds and an evaluated point with $f(\theta)<0$ is a concrete counterexample.
Validation therefore amounts to establishing that the global minimum of $f$ remains positive throughout $\Theta$.

To search a multidimensional region efficiently, \htv{} maps the unit interval onto the normalised region using a Hilbert space-filling curve $h_{N,m}:[0,1]\rightarrow\Theta$, whose construction discretises each of the $N$ dimensions at a resolution of $m$ bits~\citep{StronginSergeyev13}.
Because the mapping is surjective in the limit, minimising $f$ over $\Theta$ reduces to minimising the one-dimensional composition $\tilde{f}(x)=f(h_{N,m}(x))$ over $x\in[0,1]$, up to the finite resolution $m$ of the Hilbert approximation.
If the original objective is Lipschitz continuous, the reduced objective is H\"older continuous with exponent $1/N$, so its rate of change can be bounded as
\begin{equation}
|\tilde{f}(x)-\tilde{f}(x')|\leq H|x-x'|^{1/N},
\end{equation}
where $H$ is a H\"older constant.

\htv{} exploits this continuity by partitioning $[0,1]$ into intervals and estimating a H\"older constant for each from local and global observations~\citep{LeraSergeyev02}, which defines lower-bounding envelopes for the unexplored objective values inside each interval.
At every iteration it selects the interval with the smallest estimated lower bound, evaluates the objective at its candidate minimiser, splits the interval, and updates the estimates, directing model evaluations towards regions that may contain either a counterexample or the global minimum.
It terminates when the selected interval is shorter than a user-provided convergence threshold\footnote{\citet{ZhangKouvarosLomuscio25} refer to this as the ``optimisation budget''; here we call it the convergence threshold to clarify that smaller thresholds induce a more fine-grained search.}, then calibrates the estimated global lower bound to account for the finite Hilbert resolution and the remaining interval width.
It reports the input as robust when the calibrated lower bound is positive, and as non-robust when an evaluated objective value is negative, returning the corresponding transformation as a witness.
A witnessed counterexample is definitive, whereas the soundness of a robust conclusion depends on the H\"older estimates upper-bounding the true H\"older constant.
The next section presents \htvv{}, our margin-aware replacement for the fixed convergence threshold; the two subsequent sections instantiate the objective for VLM and VLA outputs.

%% file: sections/general_changes_to_h2v.tex
\section{\htvv{}: Margin-aware Convergence at Scale}
This section presents \htvv{}, our extension of \htv{} from sub-billion-parameter vision models to VLMs and VLAs.
The binding constraint at this scale is the cost of a query: each objective evaluation is a forward pass of a model with up to 32B parameters. 
The fixed convergence threshold used by the original implementation \citep{ZhangKouvarosLomuscio25} interacts badly with this constraint in both directions.
\begin{itemize}
    \item \textbf{Too small: wasted queries.}
    The search keeps splitting intervals whose outcome can no longer change, spending expensive queries on regions where the H\"older bound already excludes a counterexample.
    \item \textbf{Too large: undecided runs.}
    The calibration term that \htv{} subtracts after convergence for the unexplored variation over the remaining interval width can erase an otherwise positive lower bound, so the run ends undecided even when all observed margins are positive.
\end{itemize}
Choosing the threshold well therefore requires knowing the margins in advance, which is exactly what the search is trying to discover.

\htvv{} replaces the fixed threshold with a global, margin-aware convergence threshold.
We keep the notation of \citet{ZhangKouvarosLomuscio25}: $l_i$ is the lower bound of interval $i$ and $l_m$ the resulting estimate of the global lower bound, $H$ and $L$ are the present estimates of the global H\"older and Lipschitz constants, $m$ is the resolution of the Hilbert approximation, and $N$ is the dimensionality of the perturbation region.
The index $k$ counts iterations of the search, and $\varepsilon_k$ denotes the convergence threshold in force at iteration $k$.
\htv{} calibrates the global lower bound as $l_m\gets l_m-\eta$, where the calibration $\eta=\eta_{\mathrm{hilb}}+\eta_{\mathrm{opt}}$ splits into a term
$\eta_{\mathrm{hilb}} = L\cdot 2^{-(m+1)}\sqrt{N}$
for the finite resolution of the Hilbert curve and a term
$\eta_{\mathrm{opt}} = H\varepsilon_k^{1/N}$
for the unexplored variation over the remaining interval width.\footnote{The appendix records where our calibration departs from \citet{ZhangKouvarosLomuscio25}.}
The convergence threshold is clamped to $[\varepsilon_{\min},\varepsilon_{\max}]$ with $\varepsilon_{\min}$ and $\varepsilon_{\max}$ as user-specified hyperparameters.
While $l_m-\eta_{\mathrm{hilb}}\leq 0$ the search keeps $\varepsilon_k=\varepsilon_{\min}$; once $l_m-\eta_{\mathrm{hilb}}>0$, it sets
\begin{equation}
    \varepsilon_k = \min\left\{\varepsilon_{\max},\;
        \max\left\{\varepsilon_{\min},\;
        \left(\frac{l_m-\eta_{\mathrm{hilb}}}{H}\right)^{N}\right\}\right\}.
    \label{eq:adaptive-convergence}
\end{equation}
Although the threshold carries an iteration index, the rule is memoryless: every iteration recomputes $\varepsilon_k$ from the current $l_m$ and $H$. 
The adapted algorithm is described as pseudocode in Algorithm~\ref{alg:adaptive} in the appendix.
The threshold is set such that the calibrated global lower bound $l_m-\eta$ is zero at the moment of convergence.
Where the original method fixes the threshold and lets the calibration consume whatever margin it happens to consume, \htvv{} fixes the margin it is able to spend and derives the threshold from it.
As the margin is dependent on the behaviour of the model, and $N$ varies for different perturbation families, the dynamic threshold is able to generalise across both models and perturbation families, whereas a fixed threshold is not.

For a more in-depth discussion of the soundness of \htvv{}, we refer to the appendix.

%% file: sections/vlm_setup.tex
\section{Validating VLM Robustness}

\paragraph{Specification.}
We evaluate the robustness of VLMs on multiple-choice visual-question-answering tasks, where a model's clean prediction is the letter option it selects, e.g., A, B, C, or D.
For each instance, we constrain decoding to the logits corresponding to the allowed letters and only generate one token.
Let $\ell_y(\theta)$ be the logit for answer letter $y$ under perturbation $\theta$, and let $y_0$ be the letter with the highest logit on the clean image.
The specification requires that $y_0$ remain highest under every perturbation in the region.
The objective margin is
$
    f(\theta) = \ell_{y_0}(\theta) - \max_{y \neq y_0}\ell_y(\theta),
$
where the maximum runs over the other option choices for that instance.
If the margin becomes negative, the model predicts a different answer and the perturbation is adversarial.
The input is robust when the clean prediction holds throughout, with ${f(\theta)>0}$ for every $\theta \in \Theta$.

\paragraph{Dataset.}
We sample 50 instances from MMBench \citep{Liu+24} as our evaluation dataset, a multiple-choice visual-question-answering benchmark with a fixed answer set that allows direct scoring across models.
Under MMBench's ability taxonomy, our sampled instances span $32$ perception and $18$ reasoning questions, with a mix of sub-abilities in each category.
Each instance consists of an image, a question, optional hint text, and two to four answer options.
We adopt the zero-shot prompt template of \citet{Liu+24}. 
A full prompt example is shown in the appendix.

\paragraph{Models.}
We evaluate six state-of-the-art, instruction-tuned VLMs across different architectures and parameter scales, from 7B to 32B: Qwen2.5-VL-7B-Instruct and Qwen2.5-VL-32B-Instruct \citep{Bai+25}, Qwen3-VL-8B-Instruct \citep{Bai+25b}, LLaVA-v1.6-Mistral-7B \citep{Liu+24b}, InternVL3.5-8B \citep{Wang+25b}, and Gemma3-27B-it \citep{Gemma+25}.
The six models use different vision encoders, spanning CLIP (LLaVA), SigLIP (Gemma), InternViT (InternVL), and dynamic-resolution ViTs (Qwen-VL).%
\footnote{
    We maintain inputs and images in float32 throughout, but the LLaVA-v1.6 and InternVL3.5 processors cast them to uint8, which might reduce the expressiveness of small perturbation strengths.
}

\paragraph{Perturbations.}
We evaluate three visual perturbations: brightness as a shift in pixel intensity, rotation as a shift in pixel position, and their composition, a brightness shift followed by a rotation.
Brightness and rotation each act on a one-dimensional region, $\Theta_{b}$ and $\Theta_{r}$ respectively, with their composition acting on the two-dimensional product $\Theta_{b} \times \Theta_{r}$.
We write $\theta_{b}$ and $\theta_{r}$ for the brightness and rotation components of a perturbation $\theta$.

\textit{Brightness.} For a brightness radius $\rho_{b}$ the region is $\Theta_{b} = [-255\rho_{b}, +255\rho_{b}]$, applied as a uniform intensity shift $\theta_{b}$ to every pixel with the result clipped back to $[0,255]$. We report $\rho_{b}$ as a percentage of the intensity range, so $\rho_{b}=0.9$ is $90\%$.

\textit{Rotation.} For a rotation radius $\rho_{r}$ the region is $\Theta_{r} = [-\pi \rho_{r}, +\pi \rho_{r}]$, applied as a rotation by angle $\theta_{r}$ about the image centre, with bilinear interpolation for resampling and black padding for pixels drawn from outside the frame. We report $\rho_{r}$ in degrees, so $\rho_{r}=0.5$ is $90^{\circ}$.

\paragraph{Radii.}
We run validation at sets of increasing radii, $B$ for brightness and $R$ for rotation,
\[
    \begin{array}{l}
        B = \{0.5,\ 1,\ 2,\ 5,\ 10,\ 20,\ 30,\ 50,\ 70,\ 90\}\ \%, \\[2pt]
        R = \{1,\ 2,\ 4,\ 10,\ 15,\ 30,\ 45,\ 60,\ 75,\ 90\}\ {}^{\circ}.
    \end{array}
\]
The composed perturbation pairs the two sets by rank, taking $B_i$ and $R_i$ to be the $i$-th smallest radius of each, giving ten severity-matched pairs $(B_i, R_i)$ for $i=1,\dots,10$.

%% file: sections/vla_setup.tex
\section{Validating VLA Robustness}

\paragraph{Specification.}
We evaluate the robustness of VLAs on single-step action prediction, where the clean prediction is the action a model commands from an unperturbed observation: a $7$-DF (degrees of freedom) end-effector delta command of three translation terms, three rotation terms, and a gripper term whose sign selects opening or closing.
Let $a^0=(t^0,r^0,g^0)$ and $a(\theta)=(t(\theta),r(\theta),g(\theta))$ be the action for the clean and $\theta$-perturbed observations, respectively.
The specification requires that the translation stay within a tolerance $\tau_{t}$ of $t^0$, that the rotations stay within a tolerance $\tau_{r}$ of $r^0$, and that the gripper keeps the sign of $g^0$.
The evaluation objective is
\begin{equation}
    f(\theta) = \min\left\{
    1 - \frac{\| t(\theta)-t^0 \|}{\tau_{t}},\,
    1 - \frac{\| r(\theta)-r^0 \|}{\tau_{r}},\,
    \frac{g(\theta)}{g^0}
    \right\},
    \label{eq:action-margin}
\end{equation}
with $\|\cdot\|$ the Euclidean norm.
We consider a perturbation adversarial when the margin turns negative, i.e., once the commanded action leaves its translation or rotation tolerance, or the grasp direction reverses.

\paragraph{Dataset.}
We use the LIBERO dataset \citep{Liu+23}, a benchmark for knowledge transfer in lifelong robot manipulation.
LIBERO is widely used for benchmarking VLA policies, due to its consistent $7$-DF action expectations on the same Franka Panda arm.
Its four evaluation suites, LIBERO-Spatial, LIBERO-Object, LIBERO-Goal, and LIBERO-10, each contain $10$ tasks with up to $50$ expert demonstrations per task.
For each task, we randomly sample $1$ episode and $1$ timestep from that episode, giving $40$ observations used as our evaluation data points.
Each observation consists of a language instruction, the proprioceptive state of the arm, and two camera views: a fixed third-person view of the workspace, and a wrist-mounted view that moves with the end-effector.

We set the two tolerances of Equation~\ref{eq:action-margin} from the full set of demonstrations across the LIBERO suites, taking the $90$th percentile of the translation and rotation distance between consecutive demonstrated actions.
This yields $\tau_{t}=0.1529$ and $\tau_{r}=0.0264$, which translates to $7.65\,\mathrm{mm}$ of translation and $0.76^\circ$ of rotation per control step, at the benchmark's $20\,\mathrm{Hz}$ control rate.

\paragraph{Models.}
We evaluate five state-of-the-art VLA robotics policies, using their checkpoints fine-tuned on the LIBERO benchmark: OpenVLA \citep{Kim+24}, OpenVLA-OFT \citep{Kim+25b}, GR00T N1.7 \citep{Bjorck+25}, $\pi_0$ \citep{Black+25b}, and $\pi_{0.5}$ \citep{Black+25}.
The five policies use different action heads, spanning discretised action tokens (OpenVLA), a regressed chunk of eight actions (OpenVLA-OFT), and flow-matching chunks over a longer horizon (GR00T N1.7, $\pi_0$, $\pi_{0.5}$).
We fix the seed used in the flow-matching policies to ensure deterministic outputs, and we only evaluate the first action predicted by each policy.

To account for the discontinuities in OpenVLA's action output, we read its seven dimensions as the softmax-weighted expectation over the centres of its $256$ action bins.
This coincides with its original argmax decoding whenever the distribution is concentrated on a single bin.
OpenVLA is also conditioned on a single camera view, while the other four VLA models take both camera views as input.

\paragraph{Perturbations.}
The visual perturbations match the VLM setup, but each observation carries two separately mounted camera views. We consider illumination to be a property of the scene, so a single brightness shift reaches both views, and pose to belong to each mount, so rotation is drawn per view.

\textit{Dual rotation.} For a rotation radius $\rho_{r}$, each view takes its own angle from $\Theta_{r} = [-\pi \rho_{r}, +\pi \rho_{r}]$, so a perturbation is a pair in $\Theta_{r}^{2}$ whose angles need not agree.

The composed region is $\Theta_{b} \times \Theta_{r}^{2}$, so brightness gives $N=1$, dual rotation $N=2$, and their composition $N=3$.
OpenVLA takes a single view, so its rotation is scalar and only that image is perturbed.

\paragraph{Radii.}
To match the spatial sensitivity of robotics policies, we keep the rank pairing of the VLM setup and restrict rotation to smaller angles, over sets
\[
    \begin{array}{l}
        B   = \{0.5,\ 1,\ 2,\ 5,\ 10,\ 20,\ 30,\ 40,\ 50,\ 70\}\ \%,              \\[2pt]
        R^2 = \{0.125,\ 0.25,\ 0.5,\ 1,\ 2,\ 4,\ 6,\ 10,\ 15,\ 20\}\ {}^{\circ}.
    \end{array}
\]
Each radius in $R^{2}$ bounds both view angles, and rank pairing with $B$ gives ten severity-matched pairs.

%% file: sections/results.tex
\section{Results}

We report an instance as robust if \htvv{} establishes that no perturbation anywhere in the bounded region changes the specified output, conditional on the H\"older estimates.
Adversarial means \htvv{} found a concrete perturbation that violates the specification.
Instances that need more iterations than the solver budget allow are reported as unknown, which means that neither condition was established.
We measure cost as the number of model queries, the hardware-independent standard cost measure for black-box search.

We set $\varepsilon_{\min}=10^{-7}$ and $\varepsilon_{\max}=10^{-3}$.
We did not tune these values, but leave a more detailed ablation for future work.

\begin{figure*}[t]
\centering
\includegraphics[width=0.88\textwidth]{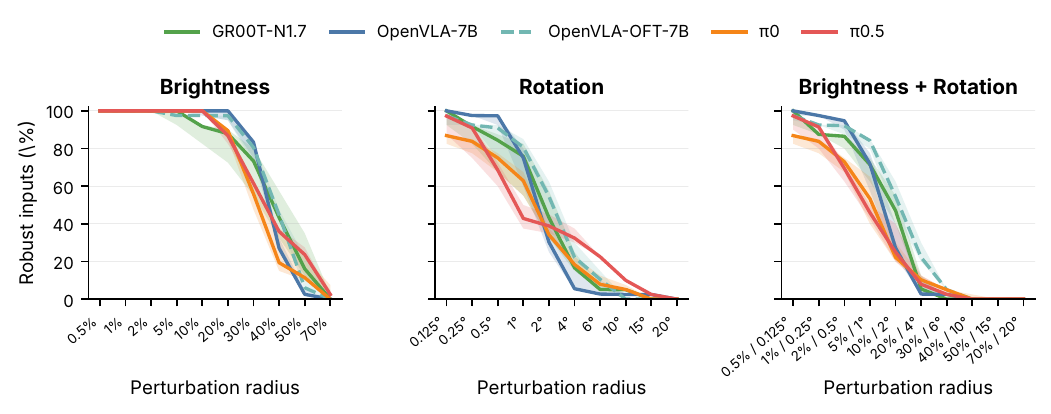}
\caption{
    VLA robustness over the $40$ LIBERO instances evaluated per model.
    Instances are unknown if they do not reach a verdict within $40\,000$ iterations.
    Lines are over the subset of instances that are robust/adversarial, bands show the uncertainty from unknown instances.
    \vspace{-0.5em}
}
\label{fig:vla_robustness}
\end{figure*}
\begin{figure*}[t]
\centering
\includegraphics[width=0.88\textwidth]{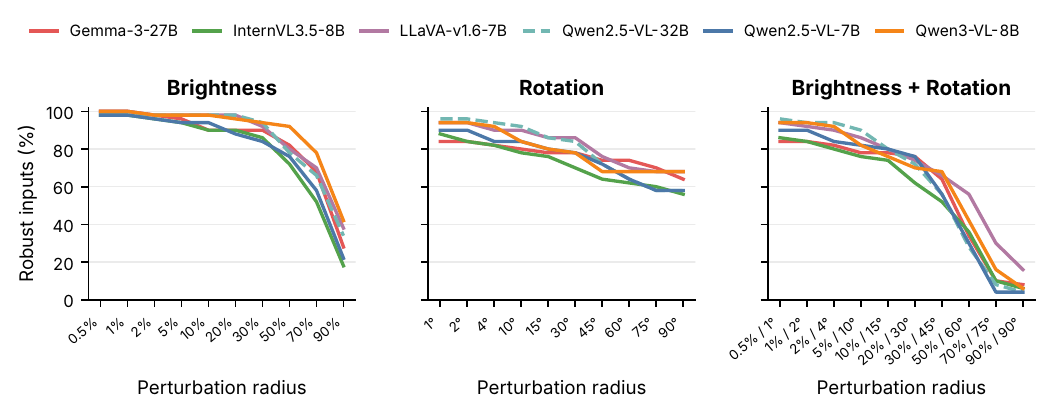}
\caption{
    VLM robustness on the 50-instance MMBench English development subset.
    \vspace{-0.5em}
}
\label{fig:vlm_robustness}
\end{figure*}

\paragraph{VLAs are more robust to brightness than to rotation.}
Figure~\ref{fig:vla_robustness} shows the validated outcomes for the five VLAs.
The models are consistently more robust to brightness than to rotation, with the combined perturbation behaving like rotation with a small additional penalty.
For all models, all $40$ sampled instances are robust up to $\pm2\%$ of the pixel range, and the rate remains $98\%$ at $\pm10\%$ and $93\%$ at $\pm20\%$.
The certificate then disappears over a narrow band, falling to $71\%$ at $\pm30\%$, and $34\%$ at $\pm40\%$.
Rotation is far more damaging, and the collapse happens at radii small enough to arise from ordinary camera mounting tolerance: aggregated over models, $96\%$ of instances are robust at $\pm0.125^\circ$, $91\%$ at $\pm0.25^\circ$, $67\%$ at $\pm1^\circ$, $40\%$ at $\pm2^\circ$, $19\%$ at $\pm4^\circ$, and $5\%$ at $\pm10^\circ$, with every instance adversarial by $\pm20^\circ$.
Crucially, a rotation of $\pm1^\circ$ or $\pm2^\circ$ is likely to occur in real-world scenarios due to remounting or vibration.

\paragraph{VLMs can tolerate larger rotations than VLAs.}
While Figure~\ref{fig:vlm_robustness} shows that VLMs are also more robust to brightness than to rotation at small radii, they remain robust to rotation far beyond the collapse of the VLA family.
This is intuitive, as the rotation of the camera directly affects the robot's understanding of the scene and the commanded action, while the VLMs are only asked to answer questions about the scene, which are likely to be rotation-invariant.

\begin{table*}[t]
\centering
\begin{tabular}{llrrrrr}
\toprule
Perturbation & Radius & GR00T-N1.7 & OpenVLA-7B & OpenVLA-OFT-7B & $\pi_0$ & $\pi_{0.5}$ \\
\midrule
Brightness & $\pm20\%$ & 88 & 100 & 97 & 89 & 87 \\
           & $\pm30\%$ & 73 & 83 & 81 & 56 & 62 \\
           & $\pm40\%$ & 43 & 27 & 45 & 19 & 36 \\
\midrule
Rotation & $\pm1^\circ$ & 76 & 75 & 81 & 63 & 43 \\
         & $\pm2^\circ$ & 43 & 30 & 55 & 34 & 39 \\
         & $\pm4^\circ$ & 17 & 6 & 22 & 18 & 32 \\
\midrule
Brightness & $\pm5\%/\pm1^\circ$ & 71 & 72 & 84 & 53 & 46 \\
$+$ Rotation & $\pm10\%/\pm2^\circ$ & 47 & 27 & 55 & 22 & 25 \\
           & $\pm20\%/\pm4^\circ$ & 5 & 3 & 22 & 10 & 8 \\
\bottomrule
\end{tabular}
\caption{
    Percentage of instances validated robust at the radii spanning each perturbation family's collapse.
    Rates condition on the instances that reached a verdict; the number of those behind each cell varies between 29 and 39 of the 40 evaluated instances.
}
\label{tab:vla_robust_rates}
\end{table*}

\paragraph{OpenVLA-OFT-7B is most robust overall, $\pi$ models are robust at large radii.}
Table~\ref{tab:vla_robust_rates} reports the per-model rates at the radii where each family collapses.
OpenVLA-OFT-7B is the most robust model overall, with the highest rate in six of the nine reported cells.
The two $\pi$ models are the weakest at the radii where the collapse begins, but are more robust than the other models at larger radii.
Aggregated over all radii and perturbations, however, the ordering is tight: OpenVLA-OFT-7B ($53.9\%$ robust), OpenVLA-7B ($50.8\%$), GR00T-N1.7 ($50.0\%$), $\pi_{0.5}$ ($48.0\%$), and $\pi_0$ ($46.2\%$).
The leader is separated by less than the width of the remaining field, and the choice of model matters far less than the choice of perturbation family or radius.
We note that the OpenVLA-7B model receives only one of two input camera images, whereas all other models, including OpenVLA-OFT-7B, receive both.
As images are independently rotated, but OpenVLA-7B is less robust than OpenVLA-OFT-7B, this suggests that the number of independent perturbations in the input does not impact robustness as much as the model's architecture and training.
Neither the flow-matching action head nor the larger action horizon of the $\pi$ models is therefore associated with greater open-loop stability; the regression-based OpenVLA-OFT head is the most stable of the five.

\paragraph{VLM robustness is defined by model family.}
Figure~\ref{fig:vlm_robustness} and Table~\ref{tab:vlm_summary} summarise the VLM evaluation.
Robustness tracks model family far more closely than parameter count: Qwen3-VL-8B and LLaVA-v1.6-7B are the two strongest models at the largest radius of every perturbation family, while the 32B member of the Qwen2.5 family improves on its 7B sibling for brightness but matches it for rotation and the combined region.
InternVL3.5-8B is the weakest model for brightness and, together with the two Qwen2.5 models, for the combined region.
Clean accuracy and robustness are close to independent: LLaVA-v1.6-7B has the lowest clean accuracy at $76\%$ yet ties for the most robust model under rotation, which is consistent with the specification measuring stability of the model's own clean answer rather than agreement with the ground truth.
\begin{table}[t]
\centering
\begin{tabular}{@{}lrrrr@{}}
\toprule
& Clean & \multicolumn{3}{c}{Robust at largest radius (\%)} \\
\cmidrule(lr){3-5}
Model & (\%) & Bright. & Rot. & Both \\
\midrule
Qwen2.5-VL-32B & 94 & 34 & 58 & 4 \\
Qwen2.5-VL-7B  & 92 & 22 & 58 & 4 \\
Qwen3-VL-8B    & 94 & 42 & 68 & 6 \\
LLaVA-v1.6-7B  & 76 & 38 & 68 & 16 \\
InternVL3.5-8B & 92 & 18 & 56 & 6 \\
Gemma3-27B     & 86 & 28 & 64 & 8 \\
\bottomrule
\end{tabular}
\caption{
    VLM clean accuracy and the percentage of instances validated robust at the largest radius of each perturbation family, over 50 MMBench instances.
    The largest radii are $\pm90\%$ brightness, $\pm90^\circ$ rotation.
}
\label{tab:vlm_summary}
\end{table}
\begin{figure}[]
\centering
\includegraphics[width=0.85\columnwidth]{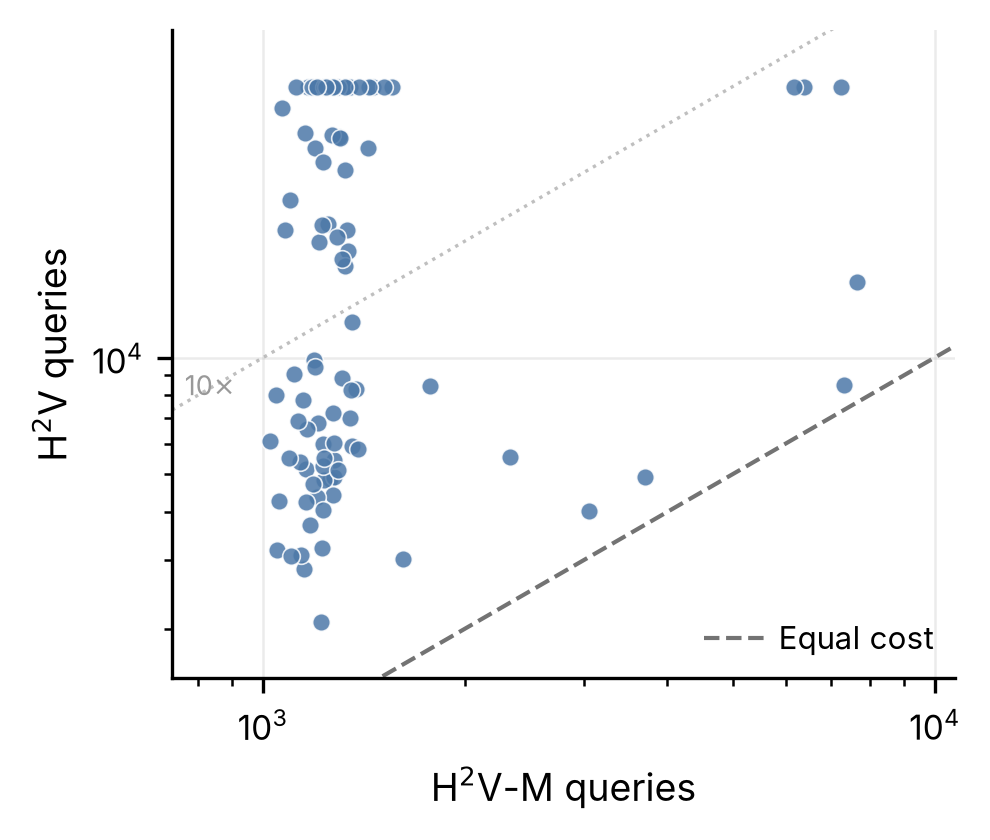}
\caption{
    Per-instance query cost of \htv{} vs. \htvv{}, over $100$ random instances \htvv{} certified within $10\,000$ queries.
    \htv{} was run with a fixed convergence threshold of $10^{-5}$ and a maximum budget of $50\,000$ queries.
    \vspace{-0.5em}
}
\label{fig:h2v_vs_h2vv}
\end{figure}
\begin{figure}[]
\centering
\includegraphics[width=0.9\columnwidth]{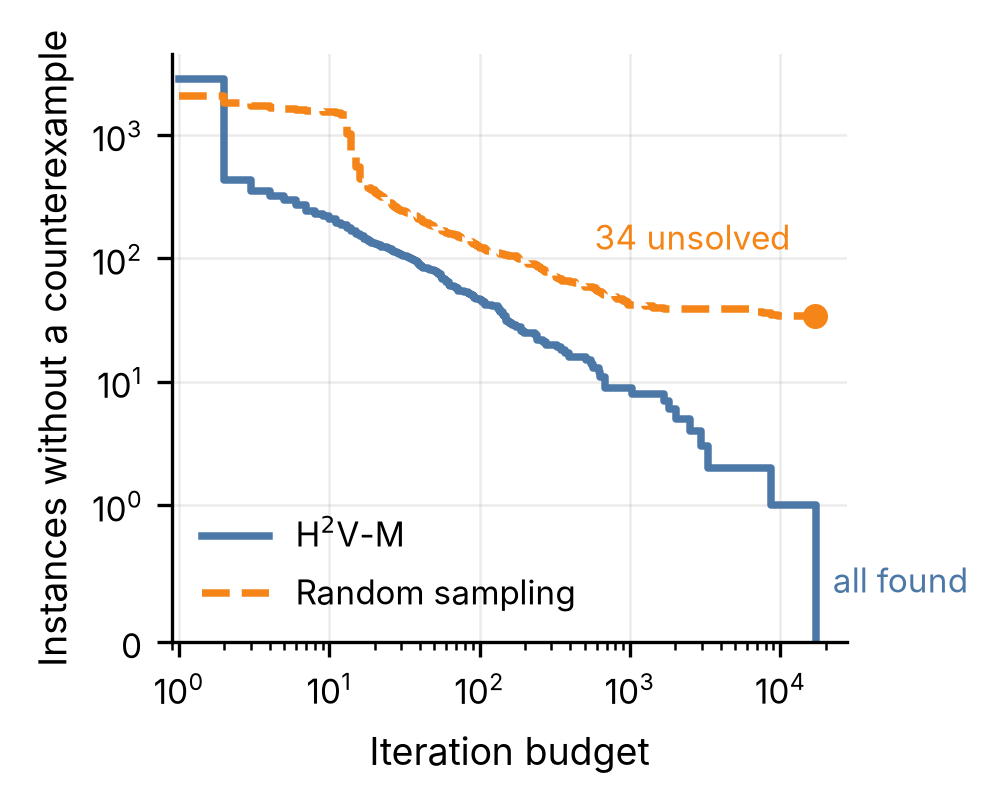}
\caption{
    Instances for which no counterexample has been found yet, as a function of the iteration budget, over the $2\,835$ instances on which \htvv{} found one.
    \htvv{} finds counterexamples faster than random sampling from.
    \vspace{-0.5em}
}
\label{fig:random_vs_h2v}
\end{figure}

\paragraph{Validation cost is strongly asymmetric.}
The VLA evaluation consumed $35.1$M model queries over $6\,000$ validation queries.
Counterexamples are cheap: the median adversarial outcome is reached in $14$ model queries, $96\%$ of them within $39$ queries, and $99\%$ within $184$.
Certificates are two orders of magnitude more expensive: the median robust outcome takes $1\,276$ queries and the mean $2\,275$.
The cost is dominated by neither: the $551$ queries that never reached a verdict consumed $29.0$M of the $35.1$M, because each runs to the iteration cap, so $9\%$ of the instances account for $83\%$ of the compute.

\paragraph{Random sampling is unsound and slower.}
For every VLA instance on which \htvv{} returned a counterexample, we draw perturbation parameters uniformly at random from the same region until one appears or $17\,000$ draws are exhausted, the number of iterations \htvv{} takes on its most expensive instance.
Neither search uses model internals or assumes transferability, so the comparison is like for like.
Figure~\ref{fig:random_vs_h2v} puts \htvv{} ahead from two iterations onward, and the gap is widest in the tail: on $34$ of the $2\,835$ instances, random sampling finds nothing within that budget.
Sampling also cannot tell those instances from ones that are genuinely robust: after $10$ or $17\,000$ unsuccessful draws the evidence is identical in kind, so the budget is a hyperparameter whose correct value is unknowable in advance and every unsuccessful run must be reported as unknown instead of robust.
\htvv{} stops only on a confirmed counterexample or a positive calibrated bound covering the whole region.

\paragraph{\htvv{} is an order of magnitude faster than \htv{}.}
We subsample $100$ instances from the VLA evaluation that \htvv{} has reported as robust within at most $10\,000$ queries, and run \htv{} on them with a maximum iteration budget of $50\,000$ queries.
\htv{} is configured to use a fixed convergence threshold of $10^{-5}$.
Figure~\ref{fig:h2v_vs_h2vv} visualises that \htvv{} requires significantly fewer queries to reach a robust verdict than \htv{}.
\htvv{} is cheaper on every one of the $100$ instances, by a median factor of $8.2$, and \htv{} fails to reach any verdict at all on $32$ of them within its budget.
\htvv{} terminated at the upper end of its threshold range, $10^{-3}$, on $90$ of the $100$ instances, two orders of magnitude coarser than a fixed $10^{-5}$.

%% file: sections/conclusion.tex
\section{Conclusion}

To our knowledge, we present the first robustness validation of state-of-the-art VLMs and VLAs over entire continuous perturbation regions, returning for each instance either a concrete counterexample or a positive validation outcome that, conditional on the estimated H\"older constants, rules out every brightness, rotation, or combined perturbation in the bounded region as a cause of output change.
Obtaining this at $32$B parameters required extending \htv{} by two orders of magnitude in model size, which the margin-aware convergence of \htvv{} delivers: on a $100$-instance subset of the VLA evaluation it decides every instance using $15.3\times$ fewer model queries in total, where the fixed threshold leaves $32$ of them undecided.

Our validation provides guarantees on the robustness of VLMs and VLAs that sampled evaluation cannot, and it shows that the conclusions drawn from sampling are often wrong.
We prove that all evaluated VLA models are robust to brightness shifts of $\pm2\%$ and that the median tolerates $\pm30\%$, whereas $27\%$ of the LIBERO instances fail under a camera rotation of $\pm1^\circ$ and $51\%$ under $\pm2^\circ$, a statement that no sampled evaluation can make.
We demonstrate that robustness is a property of the task and perturbation, not of the model: the VLMs tolerate larger camera rotations than the VLAs, and within each family the model size is less important than the choice of architecture.

Robust VLM results establish answer stability, not answer correctness; robust VLA results establish open-loop action stability, not closed-loop task success.
Both inherit the soundness caveat of \htv{}: valid overestimates of the H\"older constants give sound robust conclusions, which we empirically validate.
Future work should evaluate models with internal chain-of-thought reasoning, and extend validation to closed-loop VLA rollouts.
\pagebreak

%% file: sections/appendix.tex
\section{\htvv{} Algorithm in Pseudocode}
\begin{algorithm}[t]
\caption{\htvv{}: global margin-aware convergence}
\label{alg:adaptive}
\begin{algorithmic}[1]
\STATE Compute lower bounds for all eligible intervals and
$l_m\leftarrow\min_i l_i$
\STATE Set $\varepsilon_k\leftarrow\varepsilon_{\min}$
\IF{$l_m-\eta_{\mathrm{hilb}}>0$}
    \STATE $\varepsilon_k\leftarrow
    \min\{\varepsilon_{\max},\max\{\varepsilon_{\min},
    ((l_m-\eta_{\mathrm{hilb}})/H)^N\}\}$
\ENDIF
\STATE Select the intervals with the smallest lower bounds, as in \htv{}
\STATE Split and evaluate selected intervals wider than $\varepsilon_k$
\STATE Record convergence for selected intervals narrower than $\varepsilon_k$
\IF{the calibrated global lower bound and convergence checks pass}
    \STATE Return robust
\ELSE
    \STATE Continue splitting and model evaluation
\ENDIF
\end{algorithmic}
\end{algorithm}
Algorithm~\ref{alg:adaptive} shows the \htvv{} algorithm in pseudocode.

\section{Calibration Details}
\label{app:calibration}

We subscript the two components of $\eta$ because \citet{ZhangKouvarosLomuscio25} also use $\eta$ for the global component of the per-interval H\"older estimate.
Their $\eta_{\mathrm{opt}}$ carries a factor of one half inside the exponentiation, $H(\varepsilon/2)^{1/N}$; the implementation we build on omits it and therefore subtracts $2^{1/N}$ more than their Theorem~1 requires.
This is conservative, so it cannot turn a non-robust instance into a robust one, but it does cost decided instances, by a factor of two on the one-dimensional regions and $\sqrt{2}$ on the two-dimensional ones.
The implementation also recovers $L$ from the H\"older estimate through their relation $H=2L\sqrt{N+3}$, and evaluates $\eta_{\mathrm{hilb}}$ at the loosened constant the search trusts, which is non-decreasing over a run, rather than at the current estimate $H$ used in $\eta_{\mathrm{opt}}$.

\section{Soundness of \htvv{}}
\label{app:soundness}
The adaptive threshold cannot turn a non-robust instance into one that is incorrectly reported as robust, because the threshold governs the search schedule and not the verdict.
Reaching $\varepsilon_k$ only makes an interval count as converged, a precondition for attempting certification; the verdict is decided by the calibrated global lower bound, which \htvv{} evaluates at the actual width of the interval currently holding the smallest bound.
A positive calibrated bound is a claim about every point of every interval, including interiors that were never probed, so no counterexample remains for further splitting to find.
A fixed threshold never supplied soundness either; it supplied slack, and making it smaller makes the bound easier to satisfy without making it more valid.

\htvv{} inherits the same soundness caveat as \htv{}: the H\"older estimates must upper-bound the true variation of the objective.
If the estimated H\"older constants underestimate the true variation of the objective, the lower bounds are invalid and a robust verdict may be wrong, for the fixed and the adaptive rule alike.
If \htv{} uses a lower fixed convergence threshold than \htvv{} decides on, it has more opportunities to stumble upon a counterexample that its own bounds have wrongly excluded.
In practice, the H\"older estimates are conservative, and we empirically test the soundness of \htvv{} using SoundnessBench \citep{Zhou+25b}.
It provides instances that carry deliberately hidden counterexamples, so that any verifier reporting them as verified is demonstrably unsound.
Crucially, the SoundnessBench instances have a higher input dimensionality than our VLM and VLA regions.
As \htvv{} maps the input region to a one-dimensional Hilbert curve, a higher input dimensionality means a smaller exponent $1/N$ and therefore a steeper H\"older bound, making this benchmark particularly challenging for our method.

\section{Concrete VLM Input Instance}
\label{app:vlm-input-example}

\begin{figure}[tb]
    \centering
    \includegraphics[width=0.55\columnwidth]{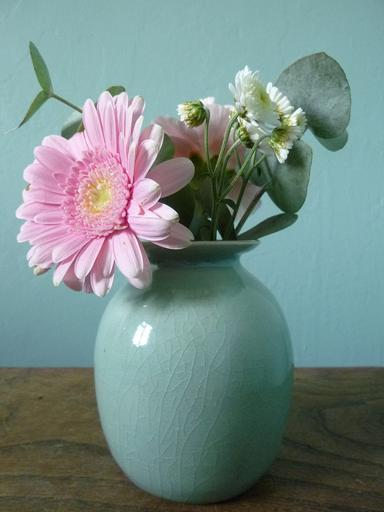}
    \caption{Image supplied with MMBench English development instance 998.}
    \label{fig:mmbench-example-998}
\end{figure}

Figure~\ref{fig:mmbench-example-998} shows instance 998 from the MMBench English development split, one of the 50 instances used in the VLM evaluation.

\paragraph{Full prompt.}
The model received the following text:
\begin{lstlisting}[numbers=none]
Question: Based on the image, what is the relation between flowers and vase?
A. Flowers are in the vase
B. Flowers are behind the vase
C. Flowers are on the top of the vase
D. Flowers are on the bottom of the vase
Answer with only one letter: A, B, C, or D.
\end{lstlisting}

\paragraph{Correct answer.}
The ground-truth answer is \textbf{A}. For its first output token, the model is expected to assign the highest logit to the character ``A'' among the four answer options. The other three answer options are treated as alternatives in the objective margin.

\section{Computing Infrastructure}
\label{app:infrastructure}

Every model is run in float32, and a run validates one instance at one radius.
The verifier runs on Python 3.12 and maps the perturbation region with \texttt{hilbertcurve} 2.0.5.
The MMBench and LIBERO evaluation instances are drawn with seed 42.

\paragraph{VLMs.}
The six VLMs were validated on NVIDIA H100 GPUs with 80\,GB of HBM3 (Hopper), served with \texttt{vllm} 0.22.0 on \texttt{transformers} 5.9.0, alongside \texttt{timm} 1.0.27, \texttt{accelerate} 1.13.0, and \texttt{qwen-vl-utils} 0.0.14.
Gemma3-27B-it and Qwen2.5-VL-32B-Instruct exceed the memory of one GPU and used tensor parallelism over two H100s, with the other four models fitting on one.
Their container is built on the \texttt{vllm/vllm-openai} v0.22.0 image, which supplies Ubuntu 22.04 with CUDA 13.0.

\paragraph{VLAs.}
OpenVLA and OpenVLA-OFT were validated on an NVIDIA A100 with 40\,GB of HBM2 (Ampere).
GR00T N1.7, $\pi_{0}$, and $\pi_{0.5}$ ran on an NVIDIA L4 with 24\,GB of GDDR6 (Ada Lovelace).
OpenVLA runs on \texttt{torch} 2.2.0, \texttt{transformers} 4.40.1, \texttt{timm} 0.9.10, \texttt{peft} 0.11.1, and \texttt{numpy} 1.26.4, and OpenVLA-OFT on the same pins with the \texttt{transformers} fork published alongside the model and \texttt{diffusers} 0.30.3.
GR00T N1.7 runs on \texttt{torch} 2.9.0, \texttt{transformers} 4.57.3, \texttt{diffusers} 0.35.1, and \texttt{numpy} 1.26.4, and both $\pi$ models on the \texttt{openpi} package published alongside them.
Their containers are built on the \texttt{nvidia/cuda} \texttt{12.8.1-devel} image for Ubuntu 24.04 LTS, which supplies CUDA 12.8; the GR00T image uses the 12.8.0 tag of the same base.
$\pi_{0}$ and $\pi_{0.5}$ take their flow-matching noise from a generator seeded at 0.
GR00T N1.7 samples inside its own policy call, and we reset the global random number generator to 0 before each inference call.

\paragraph{Experiments.}
The random-sampling comparison and the comparison against \htv{} both draw their instances from the VLA evaluation, so each runs on the hardware of the model it evaluates: OpenVLA and OpenVLA-OFT on the A100, and GR00T N1.7, $\pi_{0}$, and $\pi_{0.5}$ on the L4.
SoundnessBench runs on the L4, in a container built on the \texttt{nvidia/cuda} \texttt{12.8.1-devel} image for Ubuntu 24.04 LTS with CUDA 12.8.

%% file: bib_short.bib
@PREAMBLE{ "\newcommand{\hoek}[1]{} "}

@inproceedings{Black+25b, 
    author    = {K.~Black and N.~Brown and D.~Driess and others},
    title     = {{$\pi_0$: A Vision-Language-Action Flow Model for General Robot Control}},
    booktitle = {Proceedings of Robotics: Science and Systems},
    year      = {2025},
}

@inproceedings{Liu+23,
    author = {B.~Liu and Y.~Zhu and C.~Gao and others},
    booktitle = {Proceedings of the 36th Annual Conference on Neural Information Processing Systems (NeurIPS23)},
    pages = {44776--44791},
    publisher = {Curran Associates, Inc.},
    title = {LIBERO: Benchmarking Knowledge Transfer for Lifelong Robot Learning},
    volume = {36},
    year = {2023}
}

@inproceedings{Zhou+25c,
    author = {X.~Zhou and G.~Tie and G.~Zhang and others},
    booktitle = {Proceedings of the 39th Annual Conference on Neural Information Processing Systems (NeurIPS25)},
    pages = {127496--127523},
    publisher = {Curran Associates, Inc.},
    title = {BadVLA: Towards Backdoor Attacks on Vision-Language-Action Models via Objective-Decoupled Optimization},
    volume = {38},
    year = {2025}
}

@inproceedings{Zhao+23,
    author = {Y.~Zhao and T.~Pang and C.~Du and others},
    booktitle = {Proceedings of the 36th Annual Conference on Neural Information Processing Systems (NeurIPS23)},
    pages = {54111--54138},
    publisher = {Curran Associates, Inc.},
    title = {On Evaluating Adversarial Robustness of Large Vision-Language Models},
    volume = {36},
    year = {2023}
}

@inproceedings{FischerBaaderVechev20,
    author = {M.~Fischer and M.~Baader and M.~Vechev},
    booktitle = {Proceedings of the 33rd Annual Conference on Neural Information Processing Systems (NeurIPS20)},
    pages = {8404--8417},
    publisher = {Curran Associates, Inc.},
    title = {Certified Defense to Image Transformations via Randomized Smoothing},
    volume = {33},
    year = {2020}
}

@inproceedings{Tramer+20,
    author = {F.~Tramer and N.~Carlini and W.~Brendel and A.~Madry},
    booktitle = {Proceedings of the 33rd Annual Conference on Neural Information Processing Systems (NeurIPS20)},
    pages = {1633--1645},
    publisher = {Curran Associates, Inc.},
    title = { On Adaptive Attacks to Adversarial Example Defenses},
    volume = {33},
    year = {2020}
}

@article{Carlini+19,
    title = {On Evaluating Adversarial Robustness},
    author = {N.~Carlini and A.~Athalye and N.~Papernot and others},
    year = {2019},
    journal = {arXiv preprint arXiv:1902.06705},
}

@inproceedings{AthalyeCarliniWagner18,
    title = {Obfuscated Gradients Give a False Sense of Security: Circumventing Defenses to Adversarial Examples},
    author = {A.~Athalye and N.~Carlini and D.~Wagner},
    booktitle = {Proceedings of the 35th International Conference on Machine Learning (ICML18)},
    volume = {80},
    pages = {274--283},
    year = {2018},
    publisher = {PMLR}
}

@inproceedings{Pang+25,
    title = {Is {OpenVLA} Truly Robust? A Systematic Evaluation of Positional Robustness},
    author = {Y.~Pang and Y.~Zhao and Z.~Zhou and others},
    booktitle = {Proceedings of the 14th International Joint Conference on Natural Language Processing and the 4th Conference of the Asia-Pacific Chapter of the Association for Computational Linguistics},
    pages = {1--6},
    year = {2025},
}

@inproceedings{Wang+25c,
	title = {Exploring the Adversarial Vulnerabilities of Vision-Language-Action Models in Robotics},
	booktitle = {Proceedings of the 2025 {IEEE}/{CVF} International Conference on Computer Vision ({ICCV25})},
	author = {T.~Wang and C.~Han and J.~Liang and others},
	year = {2025},
	pages = {6948--6958},
}

@inproceedings{Zhang+25c,
	title = {Anyattack: Towards Large-scale Self-supervised Adversarial Attacks on Vision-language Models},
	booktitle = {Proceedings of the 2025 {IEEE}/{CVF} Conference on Computer Vision and Pattern Recognition ({CVPR25})},
	author = {J.~Zhang and J.~Ye and X.~Ma and others},
	year = {2025},
	pages = {19900--19909},
}

@inproceedings{Qiu+25,
	title = {Benchmarking Multimodal Large Language Models Against Image Corruptions},
	booktitle = {Proceedings of the 2025 {IEEE}/{CVF} International Conference on Computer Vision ({ICCV25})},
	author = {X.~Qiu and M.~Kan and Y.~Zhou and S.~Shan},
	year = {2025},
	pages = {9014--9023},
}

@inproceedings{Black+25,
	title = {$\pi_{0.5}$: a Vision-Language-Action Model with Open-World Generalization},
	booktitle = {Proceedings of The 9th Conference on Robot Learning (CoRL25)},
	publisher = {PMLR},
	author = {K.~Black and N.~Brown and J.~Darpinian and others},
	year = {2025},
	pages = {17--40},
}

@article{Bjorck+25,
	title = {GR00T N1: An Open Foundation Model for Generalist Humanoid Robots},
    journal = {arXiv preprint 2503.14734},
	author = {NVIDIA and J.~Bjorck and F.~Castañeda and others},
	year = {2025},
}

@article{Kim+25b,
    title = {Fine-Tuning Vision-Language-Action Models: Optimizing Speed and Success},
    author = {M.~J.~Kim and C.~Finn and P.~Liang},
    journal = {arXiv preprint arXiv:2502.19645},
    year = {2025}
}

@inproceedings{Kim+24,
	title = {OpenVLA: An Open-Source Vision-Language-Action Model},
	booktitle = {Proceedings of The 8th Conference on Robot Learning (CoRL24)},
	publisher = {PMLR},
	author = {M.~J.~Kim and K.~Pertsch and S.~Karamcheti and others},
	year = {2024},
	pages = {2679--2713},
}

@inproceedings{Zitkovich+23,
	title = {RT-2: Vision-Language-Action Models Transfer Web Knowledge to Robotic Control},
	booktitle = {Proceedings of The 7th Conference on Robot Learning (CoRL23)},
	publisher = {PMLR},
	author = {B.~Zitkovich and T.~Yu and S.~Xu and others},
	year = {2023},
	pages = {2165--2183},
}

@article{Gemma+25,
    title = {Gemma 3 Technical Report},
    journal = {arXiv preprint 2503.19786},
    author = {Gemma~Team and A.~Kamath and J.~Ferret and others},
    year = {2025},
}

@article{Wang+25b,
    title = {InternVL3.5: Advancing Open-Source Multimodal Models in Versatility, Reasoning, and Efficiency},
    journal = {arXiv preprint 2508.18265},
    author = {W.~Wang and Z.~Gao and L.~Gu and others},
    year = {2025},
}

@misc{Liu+24b,
    title = {LLaVA-NeXT: Improved reasoning, OCR, and world knowledge},
    howpublished={https://llava-vl.github.io/blog/2024-01-30-llava-next/},
    author = {H.~Liu and C.~Li and Y.~Li and others},
    year = {2024},
}

@article{Bai+25b,
	title = {Qwen3-VL Technical Report},
    journal = {arXiv preprint 2511.21631},
	author = {S.~Bai and Y.~Cai and R.~Chen and others},
	year = {2025},
}

@article{Bai+25,
	title = {Qwen2.5-VL Technical Report},
    journal = {arXiv preprint 2502.13923},
	author = {S.~Bai and K.~Chen and X.~Liu and others},
	year = {2025},
}

@article{LeraSergeyev02,
    title = {Global Minimization Algorithms for H{\"o}lder Functions},
    author = {D.~Lera and Y.~Sergeyev},
    journal = {BIT Numerical Mathematics},
    volume = {42},
    number = {1},
    pages = {119--133},
    year = {2002},
}

@book{StronginSergeyev13,
    title={Global optimization with non-convex constraints: Sequential and parallel algorithms},
    author={R.~Strongin and Y.~Sergeyev},
    year={2013},
    publisher={Springer}
}

@inproceedings{Liu+24,
    author={Y.~Liu and H.~Duan and Y.~Zhang and others},
    title={MMBench: Is Your Multi-modal Model an All-Around Player?},
    booktitle={Proceedings of the 18th European Conference on Computer Vision (ECCV24)},
    year={2024},
    publisher={Springer},
    pages={216--233},
}

@article{Qi+24,
    title={Visual Adversarial Examples Jailbreak Aligned Large Language Models},
    volume={38},
    number={19},
    journal={Proceedings of the 38th AAAI Conference on Artificial Intelligence (AAAI24)},
    author={X.~Qi and K.~Huang and A.~Panda and others},
    year={2024},
    pages={21527--21536}
}

@inproceedings{Seferis+25,
    title = {Randomized Smoothing Meets Vision-Language Models},
    author = {E.~Seferis and C.~Wu and S.~Kollias and others},
    booktitle = {Proceedings of the 2025 Conference on Empirical Methods in Natural Language Processing (EMNLP25)},
    year = {2025},
    publisher = {Association for Computational Linguistics},
    pages = {27468--27478},
}

@article{Zhou+25b,
    title={SoundnessBench: A Soundness Benchmark for Neural Network Verifiers},
    author={X.~Zhou and K.~Shen and A.~Xu and others},
    journal={Transactions on Machine Learning Research},
    year={2025},
}

@article{Kaulen+25,
    title={The 6th International Verification of Neural Networks Competition (VNN-COMP 2025): Summary and Results}, 
    author={K.~Kaulen and T.~Ladner and S.~Bak and others},
    year={2025},
    journal={arXiv preprint 2512.19007},
}

@inproceedings{Engstrom+19,
    title = 	 {Exploring the Landscape of Spatial Robustness},
    author =       {L.~Engstrom and B.~Tran and D.~Tsipras and others},
    booktitle = 	 {Proceedings of the 36th International Conference on Machine Learning (ICML19)},
    pages = 	 {1802--1811},
    year = 	 {2019},
    volume = 	 {97},
    publisher =    {PMLR},
}

@inproceedings{ZhangKouvarosLomuscio25,
    title={Scalable Neural Network Geometric Robustness Validation via H\"older Optimisation},
    author={Y.~Zhang and P.~Kouvaros and A.~Lomuscio},
    booktitle={Proceedings of the 39th Annual Conference on Neural Information Processing Systems (NeurIPS25)},
    year={2025},
    publisher={OpenReview.net},
}

@inproceedings{CohenRosenfeldKolter19,
    title = {Certified Adversarial Robustness via Randomized Smoothing},
    author = {J.~Cohen and E.~Rosenfeld and Z.~Kolter},
    booktitle = {Proceedings of the 36th International Conference on Machine Learning (ICML19)},
    pages = {1310--1320},
    year = {2019},
    publisher = {PMLR},
}

@inproceedings{Mohapatra+20,
    author = {J.~Mohapatra and T.-W.~Weng and P.-Y.~Chen and others},
    title = {Towards Verifying Robustness of Neural Networks Against A Family of Semantic Perturbations},
    booktitle = {Proceedings of the IEEE Conference on Computer Vision and
    Pattern Recognition ({CVPR}20)},
    publisher = {IEEE},
    year = {2020},
    pages = {241-249}
}

@inproceedings{Xu+21,
    title     = {Fast and Complete: Enabling Complete Neural Network
                 Verification with Rapid and Massively Parallel Incomplete
                 Verifiers},
    author    = {K.~Xu and H.~Zhang and S.~Wang and others},
    booktitle = {Proceedings of the 9th International Conference on
                 Learning Representations (ICLR21)},
    year      = {2021},
    publisher = {OpenReview.net}
}

@inproceedings{Gehr+18a,
    title		 = {{AI\textsuperscript{2}}: Safety and robustness certification of 
    neural networks with abstract interpretation},
    author		 = {T.~Gehr and M.~Mirman and D.~Drachsler-Cohen and others},
    booktitle	 = {IEEE Symposium on Security and Privacy (SP18)},
    pages		 = {3--18},
    year		 = {2018},
    organization = {IEEE}
}

@article{Liu+20a,
    title		= {Algorithms for verifying deep neural networks},
    author		= {C.~Liu and T.~Arnon and C.~Lazarus and others},
    journal		= {Foundations and Trends{\textregistered} in Optimization},
    volume		= {3-4},
    pages 		= {244-404},
    year		= {2020},
    publisher	= {Now Publishers, Inc.}
}

@incollection{Balunovic+19a,
	title         = {Certifying Geometric Robustness of Neural Networks},
	author        = {M.~Balunovic and M.~Baader and G.~Singh and others},
	booktitle     = {Proceedings of the 33rd Annual Conference on Neural Information Processing Systems (NeurIPS19)},
	pages         = {15313--15323},
	publisher     = {Curran Associates, Inc.},
	year          = {2019},
}

@article{Bunel+20,
    author =  {R.~Bunel and J.~Lu and I.~Turkaslan and others},
    title =  {Branch and Bound for Piecewise Linear Neural Network Verification},
    journal = {Journal of Machine Learning Research},
    volume = {21},
    number = {42},
    pages = {1--39},
    year = {2020},
}

@article{Singh+19a,
    title     = {An abstract domain for certifying neural networks},
    author    = {G.~Singh and T.~Gehr and M.~P{\"u}schel and
                  M.~Vechev},
    journal   = {Proceedings of the ACM on Programming Languages},
    volume    = {3},
    number    = {POPL},
    pages     = {41},
    year      = {2019},
    publisher = {ACM}
}

@inproceedings{GoodfellowShlensSzegedy15,
    title     = {Explaining and harnessing adversarial examples},
    author    = {I.~Goodfellow and J.~Shlens and C.~Szegedy},
    booktitle = {Proceedings of the 3rd International Conference on
                  Learning Representations (ICLR15)},
    year      = {2015}
}

@inproceedings{Szegedy+14,
    title     = {Intriguing properties of neural networks},
    author    = {C.~Szegedy and W.~Zaremba and I.~Sutskever and others},
    booktitle = {Proceedings of the 2nd International Conference on
                  Learning Representations (ICLR14)},
    year      = {2014}
}

@inproceedings{TjengXiaoTedrake19,
    author    = {V.~Tjeng and K.~Xiao and R.~Tedrake},
    title     = {Evaluating Robustness of Neural Networks with Mixed
                  Integer Programming},
    booktitle = {Proceedings of the 7th International Conference on 
	Learning Representations (ICLR19)},
    year      = {2019}
}

@inproceedings{Katz+17,
  author    = {G.~Katz and C.~Barrett and D.~Dill and others},
  title     = {Reluplex: An Efficient {SMT} Solver for Verifying Deep Neural Networks},
  booktitle = {Proceedings of the 29th International Conference on Computer Aided Verification (CAV17)},
  pages     = {97--117},
  year      = {2017},
  series    = {Lecture Notes in Computer Science},
  volume    = {10426},
  publisher = {Springer},
}

@inproceedings{Li+21,
    author = {L.~Li and M.~Weber and X.~Xu and others},
    title = {{TSS}: Transformation-Specific Smoothing for Robustness Certification},
    booktitle = {Proceedings of the 2021 ACM SIGSAC Conference on Computer and Communications Security (CCS21)},
    year = {2021},
    pages = {535--557},
    publisher = {ACM}
}
